\documentclass{article} 
\usepackage{nips13submit_e,times}
\usepackage{url}
\usepackage{amsmath}
\usepackage{amssymb}
\usepackage{graphicx}
\usepackage{epstopdf}
\usepackage{subfigure}

\title{Fast Training of Convolutional Networks through FFTs}

\author{
Michael Mathieu \\
Courant Institute of Mathematical Sciences\\
New York University\\
\texttt{mathieu@cs.nyu.edu} \\
\And
Mikael Henaff \\
Courant Institute of Mathematical Sciences \\
New York University \\
\texttt{mbh305@nyu.edu} \\
\And
Yann LeCun \\
Courant Institute of Mathematical Sciences \\
New York University \\
\texttt{yann@cs.nyu.edu} \\
}

\nipsfinalcopy 

\begin{document}

\maketitle

\begin{abstract}
Convolutional networks are one of the most widely employed architectures in computer vision and machine learning. 
In order to leverage their ability to learn complex functions, large amounts of data are required for training. 
Training a large convolutional network to produce state-of-the-art results can take weeks, even when using modern GPUs. 
Producing labels using a trained network can also be costly when dealing with web-scale datasets. 
In this work, we present a simple algorithm which accelerates training and inference by a significant factor, and can yield improvements of over an order of magnitude compared to existing state-of-the-art implementations. 
This is done by computing convolutions as pointwise products in the Fourier domain while reusing the same transformed feature map many times.
The algorithm is implemented on a GPU architecture and addresses a number of related challenges. 
\end{abstract}

\section{Introduction}

As computer vision and machine learning aim to solve increasingly challenging tasks, models of greater complexity are required. 
This in turn requires orders of magnitude more data to take advantage of these powerful models while avoiding overfitting. 
While early benchmark datasets in machine learning contained thousands or tens of thousands of samples \cite{LiFergusPerona-04,Bosch-07,Tzanetakis-02}, current datasets are of the order of millions \cite{Deng-2009,Bertin-Mahieux-2011}. 
This brings about new challenges as to how to train networks in a feasible amount of time. 
Even using parallel computing environments, training a network on ImageNet can take weeks \cite{KrizhevskySH12}.
In addition, although inference of labels using a trained network is comparatively fast, real-world applications such as producing labels for all images on the internet can represent a significant cost in terms of time and resources. 
Therefore, there is an important need to develop fast algorithms for training and inference.

In this work, we present a simple algorithm which accelerates training and inference using convolutional networks. 
The idea is based on performing convolutions as products in the Fourier domain, and reusing transformed feature maps many times. 
The significant operations in training convolutional networks can all be viewed as convolutions between pairs of 2-D matrices, which can represent input and output feature maps, gradients of the loss with respect to feature maps, or weight kernels.
Typically, convolutions are performed for all pairings between two sets of 2-D matrices.
By computing the Fourier transforms of the matrices in each set once, we can efficiently perform all convolutions as pairwise products. 

Although it has long been known that convolutions can be computed as products in the Fourier domain, until recently the number of feature maps used in convolutional networks has been too small to make a method like ours effective.
Previous work in the 90's \cite{Ben-Yacoub-1999} explored the possibility of using FFTs to accelerate inference at the first layer of a trained network, where the Fourier transforms of the filters could be precomputed offline. 
However, this was not used during training, possibly because the number of feature maps used at the time was too small to make the overhead of computing FFTs at every iteration worthwhile.
When the number of feature maps is large, as is the case for modern convolutional networks, using FFTs accelerates training and inference by a significant factor and can lead to a speedup of over an order of magnitude.

\section{Theory}

\subsection{Backpropagation}

The backpropagation algorithm \cite{lecun-98b} is the standard method to compute the gradient when training a convolutional network.
During training, each layer performs three tasks, which we now describe. 
First we fix some notation: for a given layer, we have a set of input feature maps $x_f$ indexed by $f$, each one being a 2-D image of dimensions $n \times n$.
The output is a set of feature maps $y_{f'}$ indexed by $f'$, which are also 2-D images whose dimension depends on the convolutional kernel and its stride. 
The layer's trainable parameters consist of a set of weights $w_{f'f}$, each of which is a small kernel of dimensions $k \times k$. 
\footnote{In this paper we assume the input images and kernels are square for simplicity of notation, but the results can be trivially extended to non-square images and kernels.}

In the forward pass, each output feature map is computed as a sum of the input feature maps convolved with the corresponding trainable weight kernel:

\begin{equation}
 y_{f'} = \sum_f x_f \ast w_{f'f}
\end{equation}

During the backward pass, the gradients with respect to the inputs are computed by convolving the transposed weight kernel with the gradients with respect to the outputs:

\begin{equation}
 \frac{\partial L}{\partial x_f} = \frac{\partial L}{\partial y_{f'}} \ast w_{f'f}^T
\end{equation}

This step is necessary for computing the gradients in (3) for the previous layer. 
Finally, the gradients of the loss with respect to the weight are computed by convolving each input feature map with the gradients with respect to the outputs: 

\begin{equation}
 \frac{\partial L}{w_{f'f}} = \frac{\partial L}{\partial y_{f'}} \ast x_f
\end{equation}
 
Note that $\frac{\partial L}{\partial y_{f'}}$ is a 2-D matrix with the same dimensions as the output feature map $y_{f'}$,
and that all operations consist of convolutions between various sets of 2-D matrices.

\subsection{Algorithm}

The well-known Convolution Theorem states that circular convolutions in the spatial domain are equivalent to pointwise products in the Fourier domain. 
Letting $\mathcal{F}$ denote the Fourier transform and $\mathcal{F}^{-1}$ its inverse, we can compute convolutions between functions $f$ and $g$ as follows:

\begin{equation*}
 f \ast g = \mathcal{F}^{-1}(\mathcal{F}(f) \cdot \mathcal{F}(g))
\end{equation*}

Typically, this method is used when the size of the convolution kernel is close to that of the input image. 
Note that a convolution of an image of size $n \times n$ with a kernel of size $k \times k$ using the direct method requires $(n-k+1)^2k^2$ operations. 
The complexity of the FFT-based method requires $6C n^2 \text{ log } n + 4n^2$ operations: each FFT requires $\mathcal{O}(n^2 \text{ log } n^2) = \mathcal{O}(2n^2 \text{ log } n) = 2Cn^2 \text{ log } n$, and the pointwise product in the frequency domain requires $4n^2$ (note that the products are between two complex numbers).
Here $C$ represents the hidden constant in the $\mathcal{O}$ notation.
\footnote{Since the FFT-based method is actually computing a circular convolution, the output is cropped to discard coefficients for which the kernel is not completely contained within the input image.
This yields an output of the same size as the direct method, and does not require additional computation.}

Our algorithm is based on the observation that in all of the operations (1), (2) and (3), each of the matrices indexed by $f$ is convolved with each of the matrices indexed by $f'$. 
We can therefore compute the FFT of each matrix once, and all pairwise convolutions can be performed as products in the frequency domain. 
Even though using the FFT-based method may be less efficient for a given convolution, we can effectively reuse our FFTs many times which more than compensates for the overhead. 

\begin{figure}
 \centering
 \includegraphics[scale=0.45]{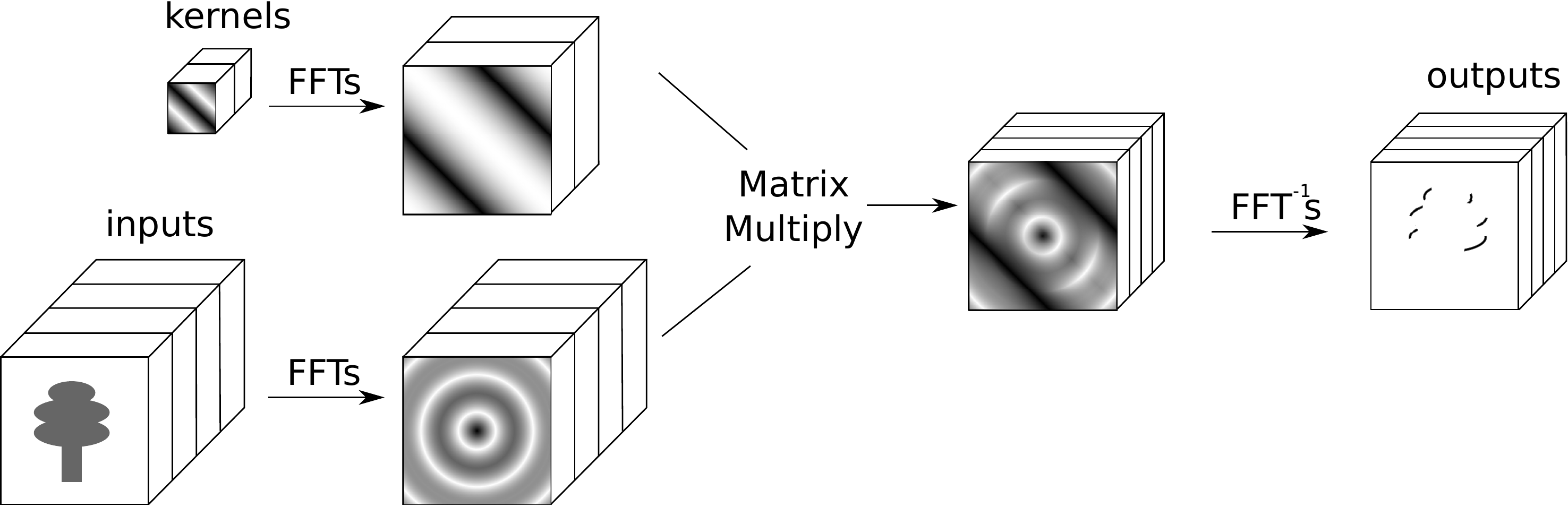}
 \caption{Illustration of the algorithm. Note that the matrix-multiplication involves multiplying all input feature maps by all corresponding kernels.}
\end{figure}

The following analysis makes this idea precise. Assume we have $f$ input feature maps, $f'$ output feature maps, images consisting of $n \times n$ pixels and kernels of $k \times k$ pixels.
Also assume we are performing updates over minibatches of size $S$, and that $C$ represents the hidden constant in the FFT complexity. As an example, using the direct approach (1) will take a total of $S \cdot f' \cdot f \cdot (n-k+1)^2 \cdot k^2$ operations. 
Our approach requires $ (2C \cdot n^2 \text{ log } n)(S \cdot f + f' \cdot f)$ operations to transform the input feature maps and kernels to the Fourier domain, 
a total of $4S \cdot f' \cdot f \cdot n^2$ additions and multiplications in the Fourier domain, and $S \cdot f' \cdot (2C \cdot n^2 \text{ log } n)$ operations to transform the output feature maps back to the spatial domain.
The same analysis yields similar complexity estimates for the other operations:

\begin{center}
  \begin{tabular}{|l| l| l| }
    \hline
    & Direct Convolution & Our Method \\
    \hline \\ [-1.5ex]
    $\sum_f x_f \ast w_{f'f}$ & $S \cdot f' \cdot f \cdot n'^2 \cdot k^2$ & $2C n^2 \text{ log } n [ f' \cdot S + f \cdot S + f' \cdot f ] + 4S \cdot f' \cdot f \cdot n^2$ \\
    \hline \\ [-1.5ex]
    $\frac{\partial L}{\partial y_{f'}} \ast w_{f'f}^T$ & $S \cdot f' \cdot f \cdot n^2 \cdot k^2$ & $2C n'^2 \text{ log } n' [ f' \cdot S + f \cdot S + f' \cdot f ] + 4S \cdot f' \cdot f \cdot n'^2$ \\
    \hline \\ [-1.5ex]   
    $\frac{\partial L}{\partial y_{f'}} \ast x_f$ & $S \cdot f' \cdot f \cdot k^2 \cdot n'^2$  & $2C n \text{ log } n^2 [ f' \cdot S + f \cdot S + f' \cdot f ] + 4S \cdot f' \cdot f \cdot n^2$ \\
    \hline
  \end{tabular}
\end{center}

Here $n'=(n-k+1)$ represents the size of the output feature map. 
Note that the high complexity of the direct method for convolution comes from the product of five terms, whereas our method has a sum of products with at most four terms.
Figure 2 shows the theoretical number of operations for direct convolution and our FFT method for various input sizes.

\subsection{Implementation and Memory Considerations}

Although conceptually straighforward, a number of challenges relating to GPU implementation needed to be addressed. 
First, current GPU implementations of the FFT such as cuFFT are designed to parallelize over individual transforms. 
This can be useful for computing a limited number of transforms on large inputs, but is not suitable for our task since we are performing many FFTs over relatively small inputs. 
Therefore, we developed a custom CUDA implementation of the Cooley-Tukey FFT algorithm \cite{Cooley-Tukey-1965} which enabled us to parallelize over feature maps, minibatches and within each 2-D transform.
Note that 2-D FFTs lend themselves naturally to parallelization since they can be decomposed into two sets of 1-D FFTs (one over rows and the other over columns), and each set can be done in parallel.

Second, additional memory is required to store the feature maps in the Fourier domain.
Note that by keeping the Fourier representations in memory for all layers after the forward pass, we could avoid recomputing several of the FFTs during the backward pass.
However, this might become prohibitively expensive in terms of memory for large networks.
Therefore we reuse the same memory for all the different convolutions in the network, so that the necessary amount of memory is determined only by the largest convolution layer.
All of the analysis in the previous section and all experiments in the remainder of the paper assume we are using this memory-efficient approach.

For a convolution layer taking an input of size $n\times n$, with $f$
input features, $f'$ output features and a minibatch of size $S$, we need to store a total of $S\cdot f + S\cdot f' + f\cdot f'$ frequency representations of size $n \times n$.
As another means to save memory, we can use symmetry properties of FFTs of real inputs to store only half the data, i.e. $n(n+1)/2$ complex numbers. 
Assuming float representations, the necessary memory in bytes is:
$$4n(n+1)(S\cdot f + S\cdot f' + f\cdot f')$$

The following table shows the amount of RAM used for typical sizes of convolutions:

\begin{center}
  \begin{tabular}{|c|c|c|c|c|}
    \hline
    $S$&$n$&$f$&$f'$&RAM used\\
    \hline
    128 & 16 & 96 & 256 & 76MB\\
    128 & 32 & 96 & 256 & 294MB\\
    64  & 64 & 96 & 256 & 784MB \\
    128 & 64 & 96 & 256 & 1159MB \\
    128 & 16 & 256 & 384 & 151MB \\
    128 & 32 & 256 & 384 & 588MB \\
    128 & 16 & 384 & 384 & 214MB \\
    128 & 32 & 384 & 384 & 830MB \\
    \hline
  \end{tabular}
\end{center}

Note that this is a relatively small additional memory requirement compared to the total amount of memory used by large networks.
 
 \begin{figure}
 \centering
 \includegraphics[scale=0.4]{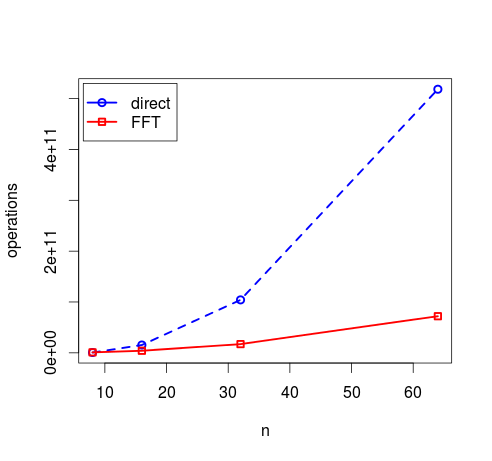}
 \caption{Number of operations required for computing (1) with different input image sizes and $S=128, f=96, f'=256, k=7$.}
\end{figure}

\section{Experiments}

To test our analysis, we ran a series of experiments comparing our method to the CudaConv GPU implementation of \cite{KrizhevskySH12} and a custom implementation using the Torch 7 machine learning environment \cite{collobert-11}.
Both of these implementations compute convolutions using the direct method in the spatial domain.
All experiments were performed on the same GeForce GTX Titan GPU. 
We began by performing unit tests comparing the results of convolutions computed by our method to those computed by the Torch implementation for each of the three operations. 
We found that the differences in results for operations (1) and (2) to be of the order of $10^{-5}$ and for operation (3) to be of the order $10^{-4}$. 
The differences are likely due to rounding errors in floating-point operations and are within an acceptable range.

We then compared how each method performed in terms of speed with varying kernel sizes, input sizes and minibatch sizes. 
The results are shown in Figure 3.
For all experiments, we chose 96 input feature maps and 256 output feature maps, which represents a typical configuration of a deep network's second layer.
The functions $\texttt{updateOutput, updateGradInput}$ and $\texttt{accGradParameters}$ correspond to the operations in (1), (2) and (3) respectively.
All times are measured in seconds.

We see that our method significantly outperforms the other two in nearly all cases. 
The improvement is especially pronounced for the $\texttt{accGradParameters}$ operation, which is the most computationally expensive. 
This is likely due to the fact that the convolution we are computing has a large kernel, for which FFTs are better suited in any case.
Also note that our method performs the same regardless of kernel size, since we pad the kernel to be the same size as the input image before applying the FFT.
This enables the use of much larger kernels, which we intend to explore in future work. 

\begin{figure}
 \centering
 \subfigure{
 \includegraphics[scale=0.5,width=\textwidth]{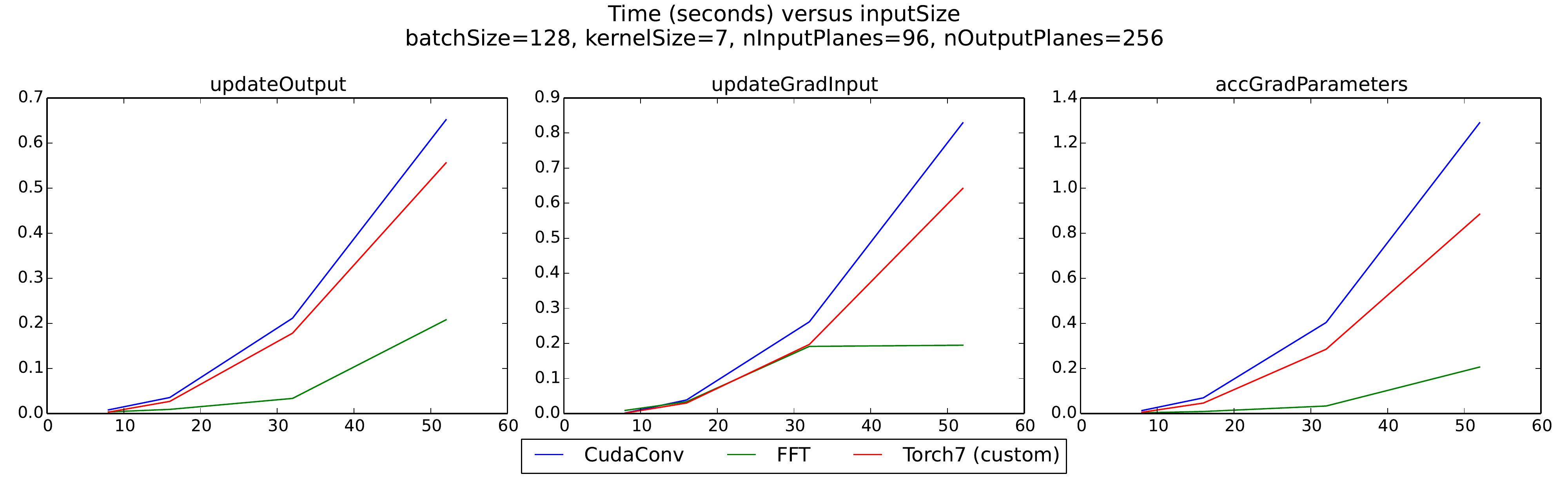}
 }
 \subfigure{
 \includegraphics[scale=0.5,width=\textwidth]{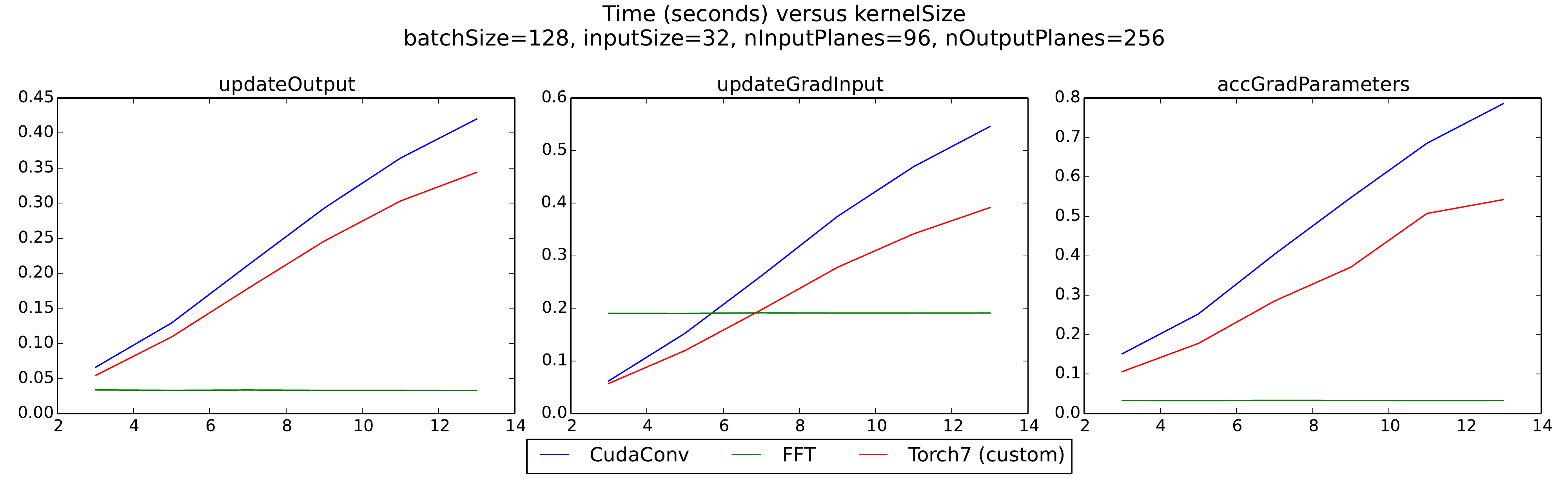}
 }
 \subfigure{
 \includegraphics[scale=0.5,width=\textwidth]{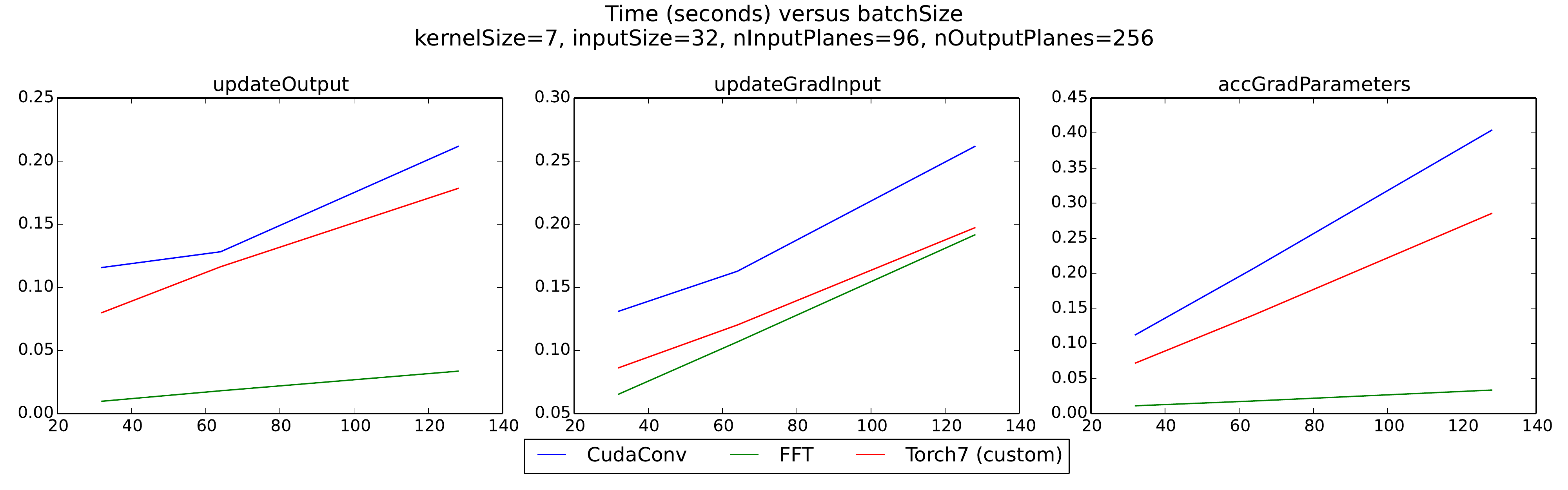}
 } 
 \caption{Speed comparison with respect to size of input image (top), kernel size (middle) and minibatch size (bottom)}
\end{figure}

\clearpage 

We next ran experiments with parameter configurations typical of those used in different layers of a large convolutional network. 
The time taken by the different methods are given in milliseconds.
The top row is a 4-tuple $(k,n,f,f')$ indicating the width of the kernel, width of the input image, number of input feature maps and number of output feature maps. 
All kernels and input images are square, of size $k \times k$ and $n \times n$ respectively. 
All configurations have minibatches of size 128.
The first configuration represents the first layer, which is why we did not report times for the \texttt{updateGradInput} operation.
For each configuration, the best-performing method is highlighted in bold.

\vspace{5mm}

\resizebox{\linewidth}{!}{%
\begin{tabular}{|c|ccccc|}
\hline
$(k,n,f,f')$ & $(11,32,3,96)$ & $(7,32,96,256)$ & $(5,16,256,384)$ & $(5,16,384,384)$& $(3,16,384,384)$ \\
\hline
\multicolumn{6}{|c|}{updateOutput} \\
\hline
Torch7 (custom) & 5 & 178 & 74 & 111 & 57 \\
CudaConv & 16 & 221 & 98 & 146 & 86 \\
FFT & \textbf{3} & \textbf{34} & \textbf{34} & \textbf{49} & \textbf{49} \\
\hline
\multicolumn{6}{|c|}{updateGradInput} \\
\hline
Torch7 (custom) & - & 197 & \textbf{76} & \textbf{116} & \textbf{62} \\
CudaConv & - & 261 & 108 & 161 & 77 \\
FFT & - & \textbf{92} & \textbf{76} & \textbf{116} & 116 \\
\hline
\multicolumn{6}{|c|}{accGradParameters} \\
\hline
Torch7 (custom) & 39 & 285 & 116 & 174 & 96 \\
CudaConv & 32 & 403 & 195 & 280 & 178 \\
FFT & \textbf{2} & \textbf{33} & \textbf{32} & \textbf{48} & \textbf{47} \\
\hline
\multicolumn{6}{|c|}{Total} \\
\hline
Torch7 (custom) & 44 & 660 & 266 & 401 & 215 \\
CudaConv & 48 & 885 & 401 & 587 & 341 \\
FFT & \textbf{5} & \textbf{159} & \textbf{142} & \textbf{213} & \textbf{212} \\
\hline
\end{tabular}}

\vspace{5mm}

We see that our FFT-based method performs faster in total for all configurations, sometimes to a substantial degree. 
The improvement is very significant on the forward pass, which makes the method especially well suited for inference on very large datasets using a trained network. 

Finally, we tested times taken to perform a training iteration for a network obtained by composing the above layers, inserting max-pooling and rectified linear units between them,  and adding a fully connected layer for prediction with 1000 outputs.
This was to account for possible changes in performance due to implementation details such as padding, accessing memory and so on.
The following table shows the results in milliseconds:

\vspace{5mm}
\begin{center}
\resizebox{0.85\linewidth}{!}{%
\begin{tabular}{|c|cccc|}
\hline
& updateOutput & updateGradInput & accGradParameters & Total  \\
\hline
Torch7 (custom) & 489 & 577 & 690 & 1756\\
CudaConv & 717 & 685 & 1093 & 2495 \\
FFT & 235 & 471 & 161 & 867 \\
\hline
\end{tabular}}
\end{center}
\vspace{5mm}

Our FFT-based method still significantly outperforms the other two implementations.

\section{Discussion and Future Work}

We have presented a simple and fast algorithm for training and inference using convolutional networks.
It outperforms known state-of-the-art implementations in terms of speed, as verified by numerical experiments. 
In the future we plan to explore the possibility of learning kernels directly in the Fourier domain.
Another interesting direction would be to investigate the use of non-linearities in the Fourier domain rather than in the spatial domain, since this would remove the need for inverse transforms and accelerate training and inference further. 

It is worth mentioning that in our current implementation of the FFT algorithm, input images which are not a power of 2 must be padded to the next highest power.
For example, using input images of size $34 \times 34$ will be suboptimal in terms of speed since they must be padded to be $64 \times 64$. 
This limitation is not intrinsic to the FFT and we intend to extend our implementation to accept other sizes in the future. 
On the other hand, the fact that our method's speed is invariant to kernel size enables us to use larger kernels at different layers of the network. 
In future work we intend to thoroughly explore the effect of input image and kernel sizes on performance.

\clearpage

\bibliographystyle{plain}
\bibliography{references.bib}

\end{document}